\documentclass[10pt,twocolumn,letterpaper]{article}

\usepackage{cvpr}      
\usepackage{booktabs} 
\usepackage{bm}
\usepackage{caption}
\usepackage[ruled]{algorithm2e} 
\usepackage{algpseudocode}
\usepackage{tabularx}
\usepackage{comment}

\definecolor{cvprblue}{rgb}{0.21,0.49,0.74}
\usepackage[pagebackref,breaklinks,colorlinks,allcolors=cvprblue]{hyperref}
\def\paperID{3871} 
\def\confName{CVPR}
\def\confYear{2026}
\begin{document}
\title{Towards High-resolution and Disentangled Reference-based Sketch Colorization}

\vspace{-2em}
\author{
*Dingkun Yan
\qquad
*Xinrui Wang\textsuperscript{1}
\qquad
*Ru Wang\textsuperscript{1}
\\
Zhuoru Li\textsuperscript{2}
\qquad
Jinze Yu\textsuperscript{3}
\qquad
Yusuke Iwasawa\textsuperscript{1}
\qquad
Yutaka Matsuo\textsuperscript{1}
\qquad
Jiaxian Guo\textsuperscript{1}
\\
\qquad
\textsuperscript{1}The University of Tokyo
\qquad
\textsuperscript{2}Project HAT
\qquad
\textsuperscript{3}Waseda University}
\vspace{-2em}
\twocolumn[{%
    \renewcommand\twocolumn[1][]{#1}
    \maketitle
    \begin{center}
        \includegraphics[width=0.98\linewidth]{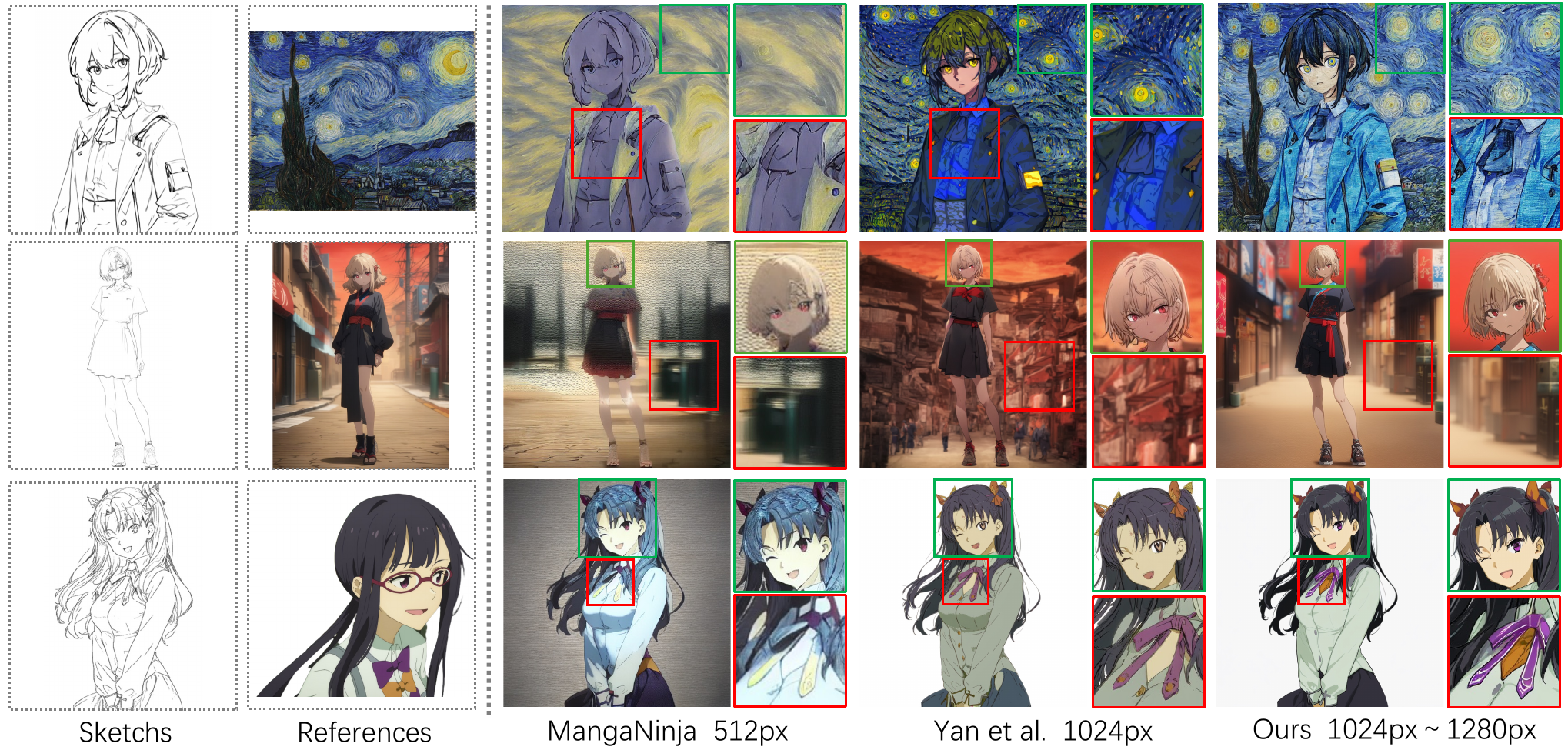}
        \vspace{-0.5em}
        \captionof{figure}{
        Left: The proposed method synthesizes colorized results in higher resolutions with accurate colors and vivid textures for inputs with various styles and contents compared to latest image-guided sketch colorization methods \cite{liu2025manganinja,yan2025image},
        Right: The proposed dual-branch architecture and gram regularization loss effectively eliminate the side effects of distribution shift.
        }
        \label{teaserfigure}
        \vspace{-0.5em}
    \end{center}%
}]
\maketitle

\def\thefootnote{\fnsymbol{footnote}}
\footnotetext[1]{Represent equal contribution to this work.}
\def\thefootnote{\arabic{footnote}}

\begin{abstract}
\vspace{-2em}

Sketch colorization is a critical task for automating and assisting in the creation of animations and digital illustrations. Previous research identified the primary difficulty as the distribution shift between semantically aligned training data and highly diverse test data, and focused on mitigating the artifacts caused by the distribution shift instead of fundamentally resolving the problem. In this paper, we present a framework that directly minimizes the distribution shift, thereby achieving superior quality, resolution, and controllability of colorization. We propose a dual-branch framework to explicitly model the data distributions of the training process and inference process with a semantic-aligned branch and a semantic-misaligned branch, respectively. A Gram Regularization Loss is applied across the feature maps of both branches, effectively enforcing cross-domain distribution coherence and stability. Furthermore, we adopt an anime-specific Tagger Network to extract fine-grained attributions from reference images and modulate SDXL's conditional encoders to ensure precise control, and a plugin module to enhance texture transfer. Quantitative and qualitative comparisons, alongside user studies, confirm that our method effectively overcomes the distribution shift challenge, establishing State-of-the-Art performance across both quality and controllability metrics. Ablation study reveals the influence of each component. Code is available: \url{https://github.com/tellurion-kanata/ColorizeDiffusionXL}.


\end{abstract}

\vspace{-1em}
\section{Introduction}
\label{sec:intro}

\begin{figure}[tb]
    \centering
    \includegraphics[width=0.97\linewidth]{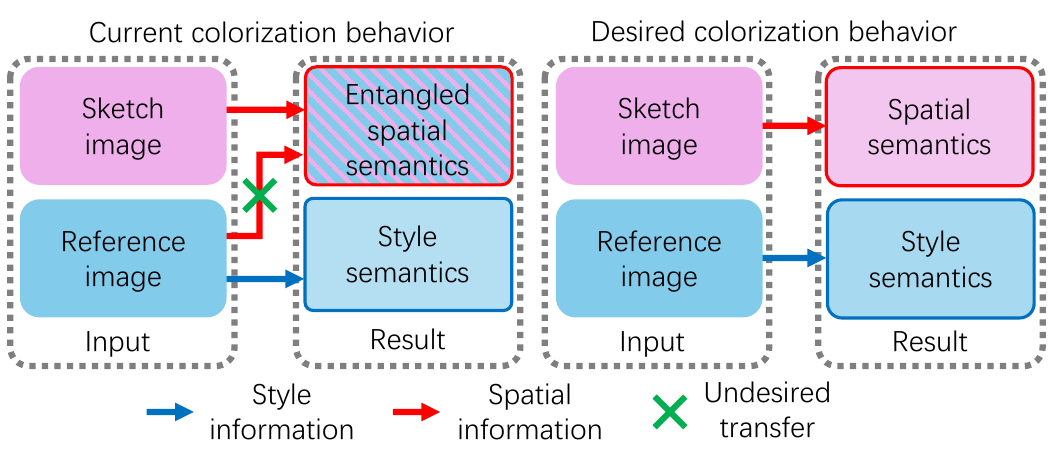}
    \vspace{-1em}
    \caption{The model incorrectly learns spatial semantics from reference images, which contradict the spatial semantics from sketch images and cause spatial entanglement.}
    \label{fig:entangle}
    \vspace{-1.5em}
\end{figure}

Animation has been a popular artistic form for decades, with the animation workflow transitioning from hand-drawn techniques with paper and celluloid to digital tools such as CLIP Studio and Adobe Animate. Recently, the integration of machine learning into animation and digital illustration workflows has gained significant traction, aiming to streamline production processes and reduce manual labor. Most methods follow the widely adopted sketch colorization paradigm, which has seen rapid advancements as the field shifted from generative adversarial network (GAN)-based approaches \cite{ZhangLW0L18, SunLWW19,zhang2017style,li2022eliminating} to the more recent diffusion-based methods \cite{animediffusion,Yan_2025_WACV,meng2024anidoc,liu2025manganinja, zhuang2025cobraefficientlineart, yan2025image, yan2025enhancing}.

The well studied image-referenced sketch colorization, mimicking professional animation production workflow \cite{yan2025image}, is fundamentally challenged by the distribution shift. This disparity exists between the semantic-aligned triplet used for training (sketch and reference derived from the ground-truth) and the potentially mismatched pairs encountered during inference. This systemic misalignment directly causes Spatial Entanglement Fig \ref{fig:entangle}, manifesting as structural contradictions in the colorized output, such as unexpected objects, color bleeding, or blurring. Previous methods, whether utilizing adjacent animation frames \cite{zhuang2025cobraefficientlineart, liu2025manganinja} or augmented ground-truth as color references \cite{yan2024colorizediffusion, yan2025image}, primarily focused on mitigating the resulting visual artifacts but failed to fundamentally resolve the distribution problem itself. Specifically, while a split cross-attention mechanism \cite{yan2025image} successfully reduced background entanglement, it proved insufficient to solve entanglement and apply precise control in the foreground regions.

\begin{figure}[tb]
    \centering
    \includegraphics[width=0.97\linewidth]{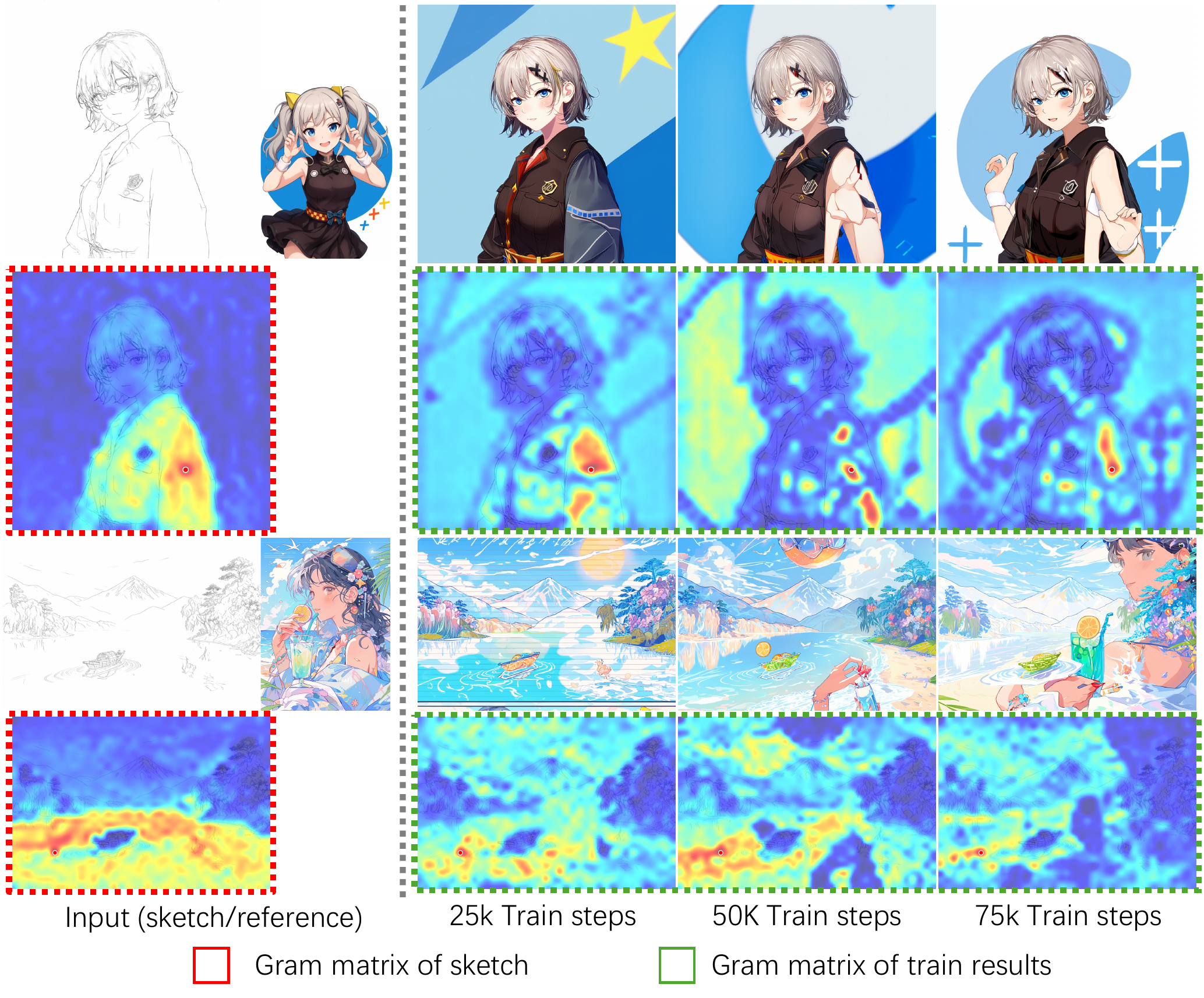}
    \vspace{-1em}
    \caption{As training progresses, the model increasingly transfers spatial semantics from the reference images into the colorized results, leading to deviations from the correct sketch-based segmentation. The ground-truth Gram matrix is obtained by discarding the reference inputs during inference, and query tokens in the Gram matrices are highlighted with red points.}
    \label{fig:dis-shift}
    \vspace{-1.5em}
\end{figure}

When analyzing the training of the colorization model, we find the model increasingly transfers spatial semantics from the reference images into the colorized results as the training processes. Shown in Fig \ref{fig:dis-shift}, it leads to the structural degradation we term spatial entanglement. To enforce that spatial information depends only on the sketch and resolve the entanglement,
We propose to explicitly model the distribution shift with a Dual-Branch Feature Alignment (DBFA) architecture, where the semantic-aligned branch models the training process and the semantic-misaligned branch imitates the inference process. A novel Gram Regularization Loss is employed on the corresponding feature maps of both branches to optimize the distribution shift by enforcing the cross-domain spatial segmentation of the diffusion backbone. To further improve colorization ability and enhance accurate control, we adopt Stable Diffusion XL (SDXL) as the diffusion backbone, with an anime-specific WD-Tagger network replacing the CLIP-L encoder for fine-grained attribute control, and a plugin module to transfer low-level visual features for improving texture synthesis and global style especially for background regions.

Extensive experiments show the remarkable abilities of the proposed method in synthesizing high-quality, high-resolution, high-controllability, and spatial-consistent colorization results. Qualitative comparisons reveal that our approach surpasses existing methods in overall image quality, detail preservation, and consistency in color distribution and geometric layouts. Quantitative evaluations also confirm its superiority over existing methods in various metrics. Moreover, in user studies, our method is consistently preferred over all the baselines.

Our contributions are as follows: 1. We propose a sketch colorization framework that explicitly models the distribution shift with two branches representing the train process and the inference process, respectively, and closes the distribution gap by a Gram regularization loss. 2. We further improve colorization quality, resolution, and controllability with an enhanced 
backbone and a novel tagger network.  3. Experiments demonstrate the effectiveness of the proposed tagger network and gram regularization loss, as well as our method's superior ability over existing methods in qualitative and quantitative comparisons, and a user study.
\vspace{-0.25em}
\section{Related work}
\label{sec:related}
\vspace{-0.25em}
\subsection{Latent Diffusion Models}
\vspace{-0.25em}


Diffusion Probabilistic Models (DPMs) \cite{HoJA20, 0011SKKEP21}, rooted in non-equilibrium thermodynamics \cite{Sohl-DicksteinW15}, have emerged as highly effective latent variable models. They have demonstrated superior performance in image synthesis and exhibit stronger conditional control compared to Generative Adversarial Networks (GANs) \cite{GoodfellowPMXWOCB14, KarrasLA19, KarrasLAHLA20, ChoiCKH0C18, ChoiUYH20}. However, the iterative denoising process, typically parameterized by a U-Net \cite{RonnebergerFB15} or Diffusion Transformer (DiT) \cite{DiT, pixart}, remains computationally intensive.

To reduce this cost, Latent Diffusion Models (LDMs), notably Stable Diffusion (SD) \cite{RombachBLEO22, sdxl}, perform diffusion and denoising within a perceptually compressed latent space using a pre-trained Variational Autoencoder (VAE). This approach reduces the computational burden. Concurrently, significant effort has been dedicated to accelerating the sampling process itself \cite{SongME21, 0011SKKEP21, 0011ZB0L022, abs-2211-01095, KarrasAAL22}. In this paper, we leverage SDXL as our core diffusion backbone. We utilize the highly efficient DPM++ solver \cite{abs-2211-01095, 0011SKKEP21, KarrasAAL22} as the default sampler, and employ Classifier-Free Guidance (CFG) \cite{DhariwalN21, ho2022classifier} to enhance the performance of our model.

\subsection{Image Referenced Diffusion Models}

Deep generative models have achieved remarkable progress in text-to-image (T2I) synthesis \cite{RombachBLEO22,sdxl,DiT,pixart}.However, fine-grained control required by real-world creative applications necessitates Image-to-Image (I2I) tasks, including image variation \cite{KwonY23, ip-adapter}, style transfer \cite{instantstyle, zhang2023inversion, ip-adapter}, and image-guided colorization \cite{animediffusion, yan2024colorizediffusion, yan2025image}. In these scenarios, reference images serve as dense visual prompts, supplying crucial details regarding color, texture, and style. The required feature extraction is task-dependent: style transfer prioritizes textural and chromatic characteristics, whereas sketch colorization demands comprehensive visual information selectively applied based on structural guidance.


Adapting pre-trained T2I architectures to I2I tasks presents a fundamental challenge due to their reliance on text-centric cross-attention. Substituting the text encoder with an image encoder for dual-conditioned I2I often leads to spatial entanglement: the reference image's semantics interfere with the target image's geometric structure (e.g., a sketch). Robustly disentangling these conflicting visual and structural feature semantics is the primary technical hurdle in developing high-quality, scalable I2I translation models.

\subsection{Sketch Colorization}


Sketch colorization has been an active research area that progressed from interactive methods \cite{SykoraDC09} to deep learning synthesis \cite{ZhangLW0L18, KimJPY19}. Current approaches fall into three guidance categories: user-guided \cite{ZhangLW0L18, s2pv5}, text-prompted \cite{KimJPY19, controlnet-iccv}, and reference-based \cite{li2022eliminating, animediffusion, yan2024colorizediffusion}. User-guided is effective but labor-intensive. Text-prompted lacks color/texture precision.


Diffusion models \cite{controlnet-v11, ip-adapter, yan2025image} advanced reference-based I2I quality. Yet, current systems remain constrained by resolution, style specificity (e.g., MangaNinja \cite{liu2025manganinja}), identity inconsistencies, and spatial artifacts. A core limitation is the failure of architectures to scale to high-resolution outputs with granular controllability.


This failure stems from Spatial Entanglement, rooted in the distribution shift between aligned training data and mismatched inference pairs. Entanglement corrupts sketch geometry, acutely worsening at high resolution and strong guidance. We eliminate this with a dual-branch framework that models the distribution shift. We leverage a Gram regularization loss to close the training/inference gap and integrate a Tagger Network for fine-grained attribution control.

\section{Method}
\label{sec:method}

\begin{figure*}[t]
    \centering
    \includegraphics[width=0.98\linewidth]{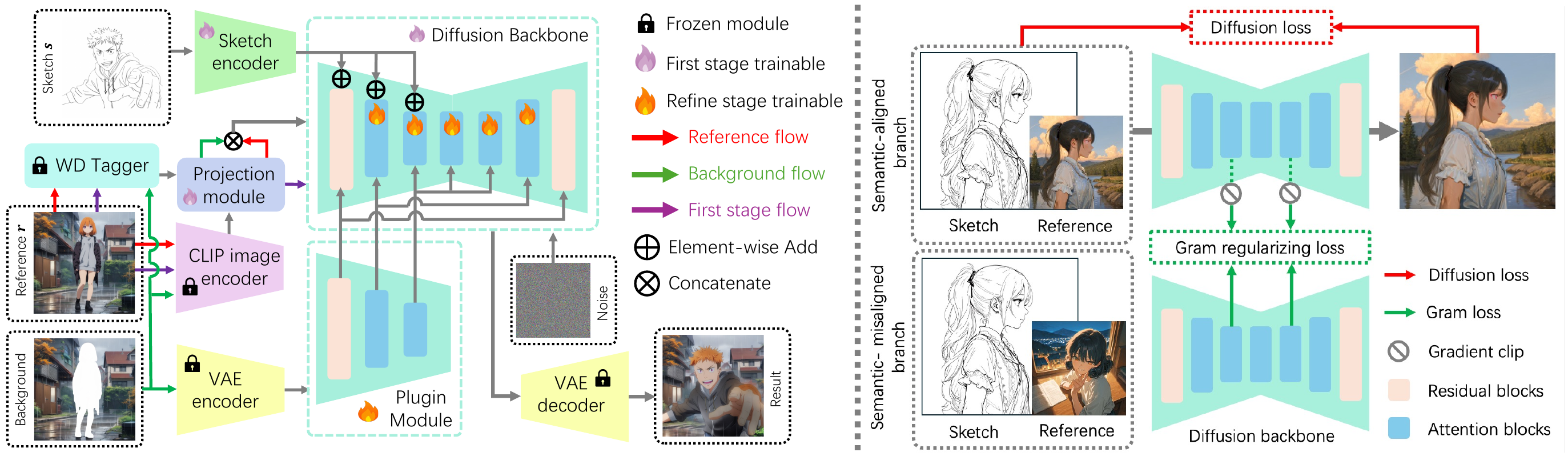}
    \vspace{-1em}
    \caption{The left panel illustrates the architecture of the proposed framework, while the right panel shows the computation of the Gram loss. In the first stage, the backbone is trained for reference-based colorization using image embeddings, where the embedding inputs to the denoising U-Net are extracted from the entire reference image (indicated by the red arrow). The Gram loss is activated only during this first training stage. In the subsequent stages, we introduce feature-level representations for foreground and background regions through their respective plugin adapters. During inference, the plugin adapters are executed only once at timestep $t=0$.}
    \label{framework}
    \vspace{-1em}
\end{figure*}



In this section, we first formalize the problem of spatial entanglement as a conditional independence failure stemming from the distribution shift. To resolve this, we introduce the Dual-Branch Feature Alignment (DBFA) architecture, which explicitly models this distribution shift during training. We then enforce structural independence using a Gram Regularization Loss, which robustly penalizes spurious spatial correlations and ensures feature-level invariance to the reference image. Finally, we integrate an anime-specific tagger network to enhance fine-grained semantic control. The overall framework is illustrated in Figure $\text{\ref{framework}}$.

\subsection{Distribution Shift and Spatial Entanglement} 
\label{sec:problem}
A core challenge in reference-based generation is the distribution shift between the training and inference phases \cite{yan2024colorizediffusion, yan2025image}. The model is trained on a distribution $P_{\text{train}}(I_r, I_s, I_{gt})$, which consists of semantically-aligned triplets where the sketch $I_s$ and reference $I_r$ are both derived from the same ground-truth image $I_{gt}$. However, at inference, the model must generalize to a shifted distribution $P_{\text{test}}(I^{\prime}_r, I^{\prime}_s)$, where the sketch and reference may be arbitrarily paired and entirely irrelevant. This mismatch incentivizes the model to learn a spurious correlation: it erroneously learns that the reference $I_r$ is predictive of the output's spatial structure, $X_{\text{spatial}}$. We formally define this structural degradation as spatial entanglement. Ideally, the spatial features $X_{\text{spatial}}$ should be conditional only on the input sketch $I_s$. The entangled state is thus: $\mathit{P(X_{\text{spatial}} \mid I_r, I_s) \neq P(X_{\text{spatial}} \mid I_s)}.\space$  This reliance on $I_r$ for spatial cues is the root cause of severe artifacts during inference, such as redundant objects, distorted body parts, and color regions that bleed across structural boundaries. As visualized in Figure~\ref{fig:dis-shift}, this entanglement can worsen as training progresses, as the model overfits to the spurious correlations in $P_{\text{train}}$. Our theoretical goal is to break this dependency and enforce spatial independence, restoring the correct conditional probability: $\mathit{P(X_{\text{spatial}} \mid I_r, I_s) = P(X_{\text{spatial}} \mid I_s)}.\space$ Achieving this disentanglement is the central motivation for our method.

\subsection{Optimize the Distribution Shift with Gram Loss}

We propose a Dual-Branch Feature Alignment (DBFA) architecture, which explicitly models the distribution shift with two weight-sharing branches to resolve spatial entanglement:
1. A semantic-aligned branch models the training process that takes sketch $I_s$ and reference $I_r$ derived from the ground truth $I_{gt}$ as input. 2. A semantic-misaligned branch models the inference process that takes randomly sampled sketch $I^{\prime}_s$ and reference $I^{\prime}_r$ pairs as input. 


Following DINOv3 \cite{dinov3}, we notice that the Gram matrix of a feature map captures the spatial correlation between different patches at the semantic level with the attention mechanism. Since spatial entanglement is driven by erroneous semantic transfer from the reference branch, we design a novel \textbf{Gram Regularization loss} which restricts the spatial correlations of internal features between two branches to enforce the semantic-misaligned branch to maintain the spatial segmentation of sketch images and eliminate the artifacts caused by the distribution shift. For computational efficiency, the loss is computed only on the features from the final transformer blocks of the U-Net's encoder and decoder at the lowest resolution.


A key distinction of our method is its ``self-anchoring'' mechanism, which operates within a single training step without needing an external network like VGG~\cite{simonyan2015deepconvolutionalnetworkslargescale} or an older model checkpoint \cite{dinov3}. For a given sketch and a noisy latent $z_t$, we perform two forward passes that differ only in the reference image provided: the \textbf{Semantic-Aligned Branch} that uses the color reference derived from ground truth, and the \textbf{Semantic-Misaligned Branch} that uses a randomly sampled reference image from the dataset.
This mechanism compels the Gram matrix of the semantic-misaligned branch to align with that of the semantic-aligned branch. Since both branches share the identical input sketch, enforcing this feature-level consistency mandates that the generated spatial features remain invariant to any color reference. It rigorously forces the network to derive structural and segmentation information exclusively from the sketch, thereby achieving the disentanglement of geometry from style. The Gram regularization loss is as follows:

\begin{equation}
\begin{tiny}
\mathcal{L}_{\text{gram}} = \sum_{l \in L} \| \text{stop\_grad}(G(x_{aligned}^{(l)})) - G(x_{misaligned}^{(l)}) \|_{F}^2
\end{tiny}
\end{equation}

where $G(x)=xx^{\top}$ is the gram matrix of feature map $x$, the set $L$ contains the indices of the targeted layers. $x_{aligned}^{(l)}$ and $x_{misaligned}^{(l)}$ are the layer-$l$ feature maps from the semantic-aligned branch and the semantic-misaligned branch, respectively. The Frobenius norm $\| \cdot \|_{F}^2$ measures the discrepancy. The \text{stop\_grad} operation detaches the anchor Gram matrix so that only the misaligned branch receives gradient updates. This prevents the anchor from drifting and collapsing toward the misaligned representation, stabilizing the optimization and ensuring that the misaligned branch consistently aligns to a fixed reference.

\begin{equation}
\mathcal{L}_{\text{diff}} = \mathbb{E}_{\mathcal{E}(y), \epsilon, t, s, c} [\|\epsilon - \epsilon_{\theta}(z_{t}, t, s, c)\|^{2}_{2}]
\end{equation}

The final training objective is defined as a weighted sum:

\begin{equation}
\mathcal{L} = \mathcal{L}_{\text{diff}} + \lambda \mathcal{L}_{\text{gram}}.
\end{equation}

We activate the Gram loss after the first 33\% of training steps ($\lambda{=}0 \rightarrow 1$) since early entanglement is minimal (Figure~\ref{fig:dis-shift}). As it is computed on two layers with gradients only through the semantic-misaligned branch, training slows by ~30\% with 10\% extra memory. See Section~\ref{sec:experiment} for full settings.

\subsection{Precise Attribution Control by WD-Tagger}

We choose Stable Diffusion XL (SDXL) as the backbone, with the weight initialized from AnimagineXL ~\cite{animagineXL} for its high-resolution synthesis capabilities. The original SDXL employs dual text encoders (OpenCLIP-bigG and CLIP-L), which often exhibit redundant semantic overlap and shared stylistic biases.

To achieve precise and style-aware control, we leverage the domain-specialized WD-Tagger network~\cite{SmilingWolf2025WD} as a replacement for the generic CLIP-L text encoder. The WD-Tagger, built upon the Swin Transformer v2 architecture ~\cite{liu2021swin, liu2022swin}, is pre-trained on a large-scale anime image dataset for multi-label classification. Compared to conventional CLIP embeddings, the resulting WD embeddings provide a more detailed and accurate representation of anime-specific attributes such as hair color, clothing type, and background theme, thereby offering more expressive control signals during the diffusion process. Furthermore, by explicitly projecting visual features into tag-aligned embeddings, the WD-Tagger ensures that the learned representations possess strong semantic grounding at the attribution level. This design facilitates robust feature clustering within the latent space. Consequently, the diffusion backbone's capacity to capture semantics from input sketches is significantly improved, leading to enhanced consistency and fidelity in the synthesized results.

We substitute the OpenCLIP-bigG text encoder with its image encoder to facilitate the extraction of image-based embeddings. Given that CLIP is inherently designed to project both text and images into a shared latent space for cosine-similarity computation, the image encoder provides lower-level visual representations that better support cross-style generalization and transfer compared to the tag-aligned embeddings of WD-Tagger. This dual-encoding design, combining the categorical control of WD-Tagger with the broad visual embeddings of OpenCLIP, furnishes the diffusion backbone with a comprehensive set of control signals, enabling high-quality, style-consistent synthesis.

\subsection{Feature-level Plugin}

The embedding-level reference inject often lacks fine-grained details, leading to inconsistent results with poor textures, particularly in background regions. The proposed Gram regularization loss may also increase the randomness of backgrounds when the reference image lacks explicit background content, leading the model to generate arbitrary or inconsistent backgrounds. To address this issue, we introduce an independent encoder as a plugin module for the refining stage to enhance the backgrounds and global style. This module learns feature-level representations for non-sketch regions and facilitates the transfer of global style features. 
The plugin module is trained with a multi-step strategy. In the first stage, we train the backbone with the DBFA and gram loss, and in the refinement stage, we optimize the plugin module and the split cross attention \cite{yan2025image} in the backbone with all the other parameters fixed. More details are included in the supplementary material.

\begin{figure}[t]
    \centering
    \includegraphics[width=0.99\linewidth]{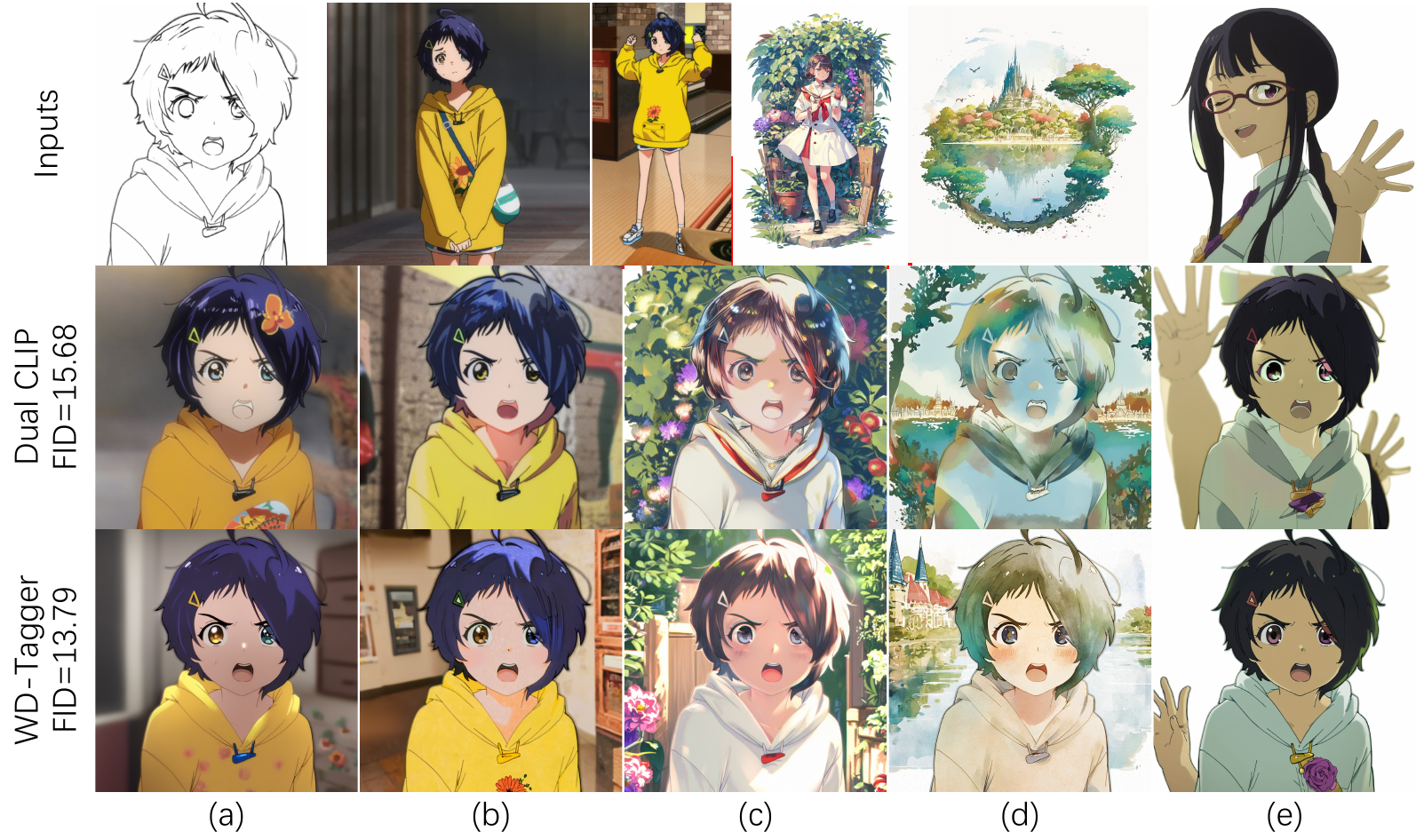}
    \vspace{-1em}
    \caption{Ablation results of WD-tagger. The model without the WD tagger fails to correctly colorize the eyes when reference eyes are small and color mismatched. It also shows weaker segmentation guidance overall. Both ablation variants exhibit artifacts without Gram regularization in (e). FID scores are shown on the left.}
    \label{fig:embedder-selection}
    \vspace{-1em}
\end{figure}

\begin{figure*}[t]
    \centering
    \includegraphics[width=0.98\linewidth]
    {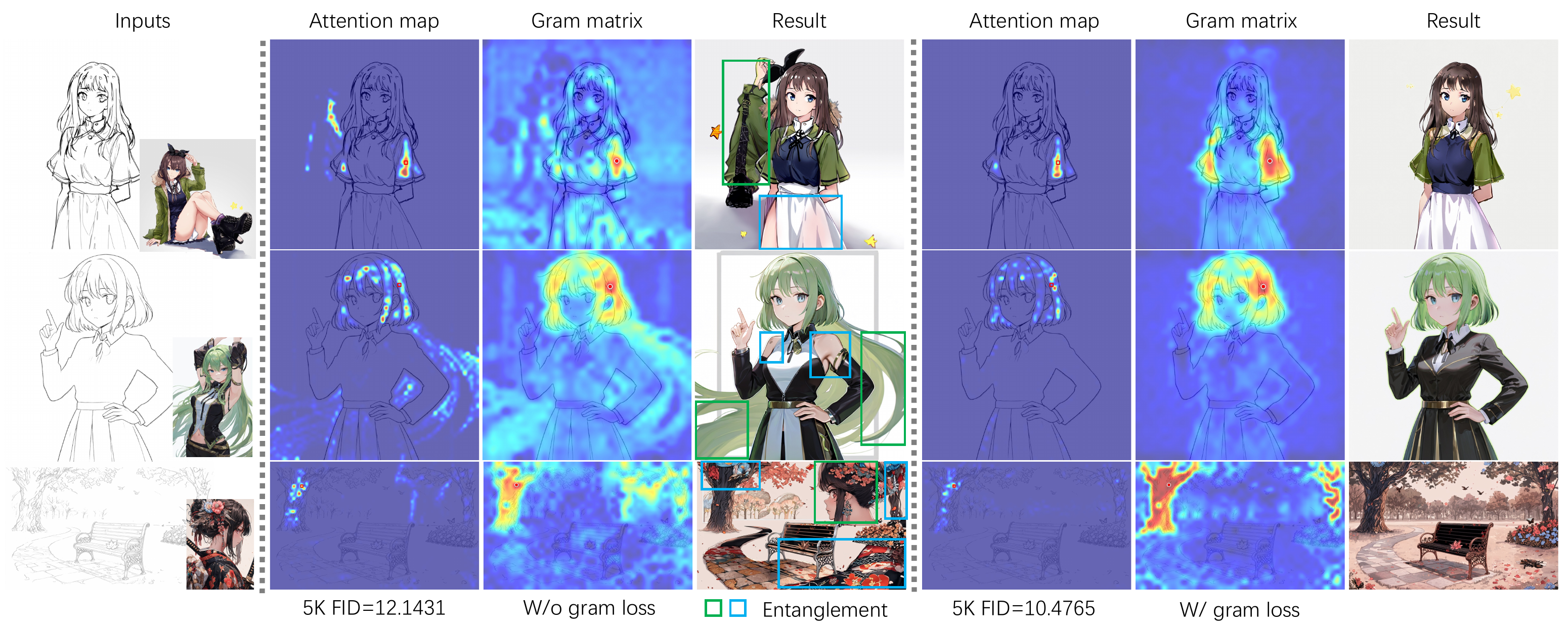}
    \vspace{-1em}
    \caption{Visualization of gram matrices and attention maps. The proposed Gram regularization loss helps enhance semantic fidelity to the sketch inputs. Query tokens are highlighted by red rectangles in both the attention maps and Gram matrices. In the generated results, green boxes highlight entanglement artifacts outside sketch regions, while blue boxes indicate semantic errors within sketch-guided areas.}
\label{fig:representation-vis}
\vspace{-1em}
\end{figure*}

\section{Experiment}
\label{sec:experiment}
\subsection{Implementation detials}
The model was trained on 8×H100 HBM3 GPUs (80GB) using DeepSpeed ZeRO-2~\cite{deepspeed}, with a total batch size of 128 and a learning rate of 1e-5. The backbone is trained for 70K steps, and the plugin module is trained for 10K steps; the full training takes ~72 hours. The training dataset focuses on high-resolution illustrations of characters and scenery, containing 6M images. To construct sketch inputs, we extracted four types of sketch representations by jointly applying edge and line extractors from~\cite{sketchKeras,xiang2022adversarial,mangaline}. 


\begin{figure}
    \centering
    \includegraphics[width=1\linewidth]{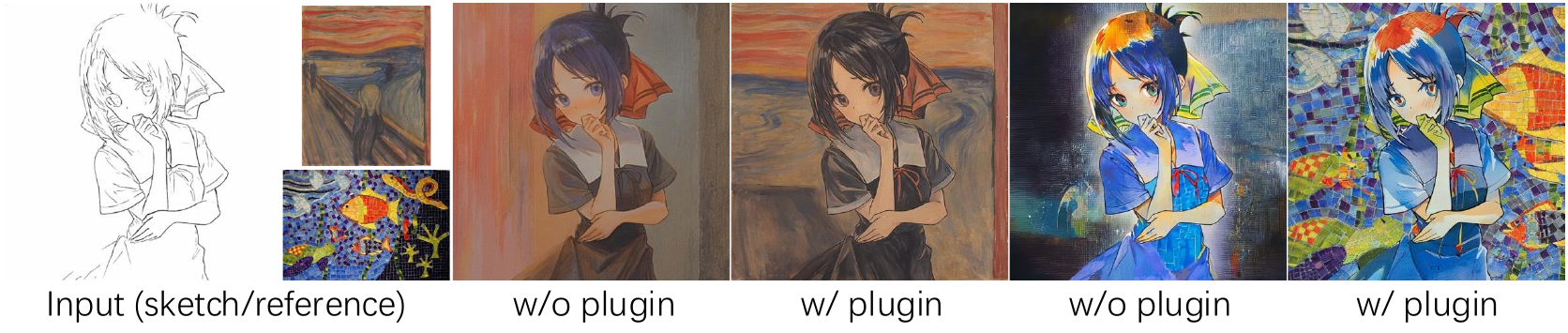}
    \vspace{-1em}
    \caption{The plugin module can be activated to inject low-level features for a higher style and background similarity.}
    \label{fig:plugin}
    \vspace{-1em}
\end{figure}

\begin{figure*}[t]
    \vspace{-1em}
    \centering
    \includegraphics[width=1\linewidth]{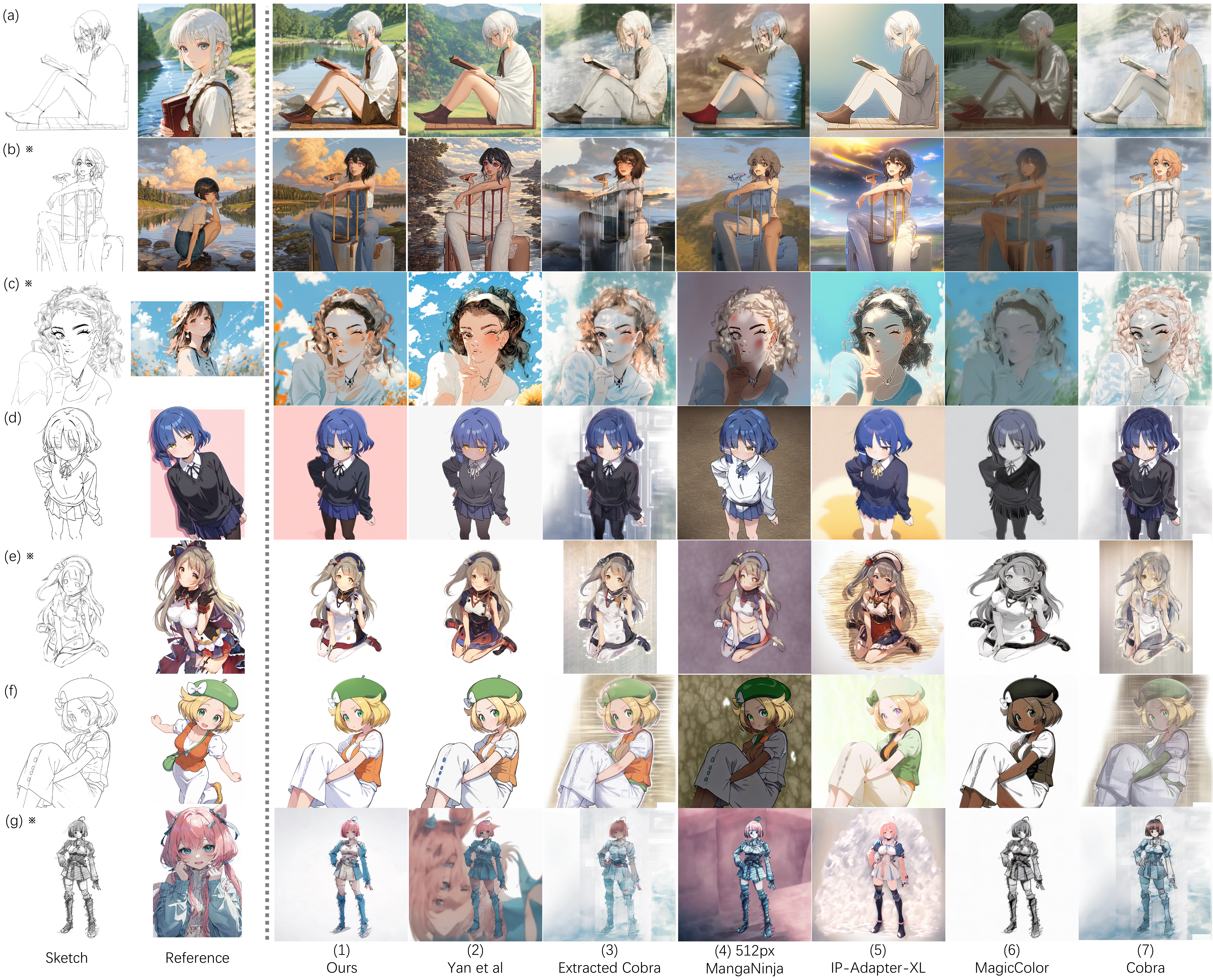}
    \caption{Qualitative comparison regarding character illustration colorization. All images are generated at $1024^{2}$ resolution except for MangaNinja \cite{liu2025manganinja}, which is fixed at $512^{2}$ in the official implementation. Zoom in for details. High-resolution images are available in the supplementary materials. Column (3) is synthesized using sketches extracted from our results. ※: Real sketches from human artists.}
    \label{fig:compairson}
    \vspace{-1em}
\end{figure*}

\subsection{Ablation study}

We perform a systematic, incremental ablation study to assess the individual contribution of each proposed component. Our baseline is an SDXL-style architecture utilizing dual OpenCLIP encoders and trained solely with diffusion loss. The proposed components are integrated cumulatively in the following two stages: 1. We replace the OpenCLIP text encoder with WD-Tagger to evaluate gains in attribute control and segmentation guidance. 2. We integrate the Gram regularization loss to confirm the promotion of spatial semantic disentanglement within the latent feature space. 3. We add a plugin module to show its effectiveness in transferring details and maintaining style consistency.


\textbf{WD tagger.} 
We exclude the Gram loss during the training of both ablation models to reveal the effect of WD Tagger, as its disentangling property suppresses embedding clustering induced by WD Tagger, thereby hindering the observation of its improvement. Our framework employs image embeddings to transfer reference information for sketch colorization. In this validation, we demonstrate that the WD Tagger provides superior embeddings compared to CLIP, yielding results that better preserve reference attributes such as eye color and texture fidelity, as shown in Figure~\ref{fig:embedder-selection}. The improvement arises as the WD Tagger is trained for multi-class classification on anime-style images, producing features that are semantically closer to text notions and can accurately capture reference information. 

\textbf{Gram loss.} The precise control signals provided by the WD tagger result in severe entanglement, while the proposed Gram loss regularizes the spatial semantics of reference-based colorization results. To highlight this improvement, we visualize the attention map of self-attention and the Gram matrices of hidden representations within the denoising U-Net in Figure~\ref{fig:representation-vis}. Both the attention maps and the Gram matrices show that the Gram loss effectively removes the entanglement of sketch semantics, preventing semantic shifts within sketch-guided regions and artifacts outside the sketches. We also report FID scores in the figure.

\textbf{Low-level plugin.} We show the qualitative comparison of the plugin module in Figure~\ref{fig:plugin}, where the results with the module show finer details and better style consistency with reference images compared to the results without the module. This validates its effectiveness in injecting low-level features to improve fine textures and details, as well as enhancing the global style. 

\subsection{Comparison with baselines}
To demonstrate the improvements achieved by our proposed framework, we conduct comparisons with several recent state-of-the-art sketch colorization methods, including ColorizeDiffusion~\cite{yan2024colorizediffusion}, Yan et al.~\cite{yan2025image}, IP-Adapter~\cite{ip-adapter}, MagicColor~\cite{peng2025exploring}, MangaNinja~\cite{liu2025manganinja}, and Cobra~\cite{zhuang2025cobraefficientlineart}.
These baselines are representative approaches that have demonstrated strong performance in transferring not only visual styles but also high-level semantic correspondences from reference images. For fair comparison, we perform evaluations at a resolution of $1024^{2}$, except for MangaNinja~\cite{liu2025manganinja}, whose official implementation supports only $512^{2}$. All baseline results are generated using their official code and pretrained weights, and the plugin module is disabled.\\

\begin{figure*}
    \centering
    \includegraphics[width=0.98\linewidth]{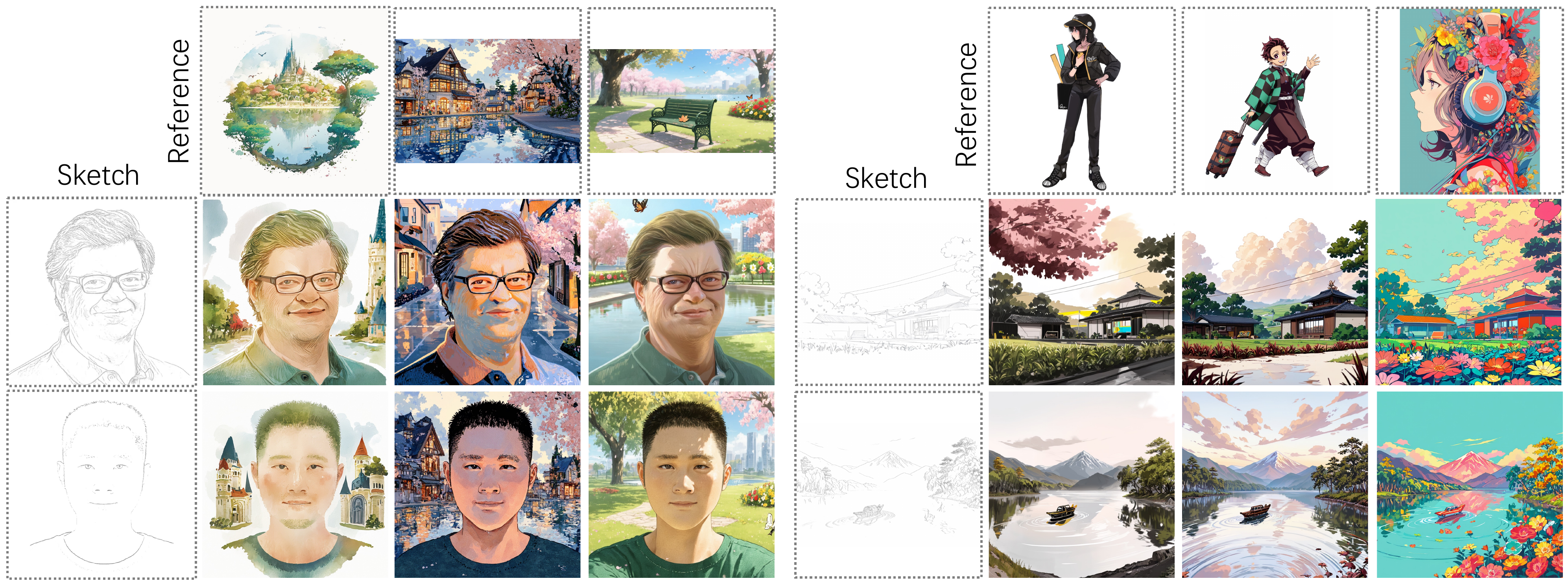}
    \vspace{-0.5em}
    \caption{We show the cross-content results, where the sketches and reference images are from different domains (for example, portrait and scenery). The proposed method effectively synthesizes visually pleasant images without artifacts.}
    \label{fig:cross-content}
    \vspace{-0.5em}
\end{figure*}

\noindent\textbf{Qualitative comparison.} 
We generate reference-based results with various pairs of inputs and visualize the result in Figure \ref{fig:compairson}. Adapter-based methods fail to synthesize results with visually pleasant textures and color distribution similar to references, and also cause artifacts in the backgrounds at such high-resolution. This is due to their inadequate generation ability. ColorizeDiffusion v1.5 \cite{yan2025image} successfully prevents artifacts and synthesizes rich textures, but the color similarity and texture quality are still less satisfying. MangaNinja \cite{liu2025manganinja} is designed for character colorization with a clear background, with resolution fixed at $512^{2}$, and trained on chopped animation frames data, making it less effective for colorization tasks with complex background textures at high resolution. Cobra shows a significant deterioration due to the change of sketch style, so we additionally generate a set of results using sketches extracted from our results, shown in column (3).

Our proposed method, on the contrary, synthesize high-quality, high-resolution, and artifact-free results characterized by fine textures and harmonious color distributions. Notably, the generated outputs demonstrate precise control over disentangled attributes, such as the rainbow in the background in (b), the color of the hat in (c), the color consistency of the sky in (d), and the saturation levels in (e). This comparison clearly validates the superior performance of the proposed method over existing approaches in terms of texture richness, color preservation, attribute disentanglement and controllability, and overall visual quality.


\noindent\textbf{Quantitative comparison.}
We report quantitative results using FID, MS-SSIM, PSNR, and CLIPScore, with FID as our primary metric due to its strong correlation with perceptual quality and its distribution-level comparison that does not require semantic or spatial alignment. Fréchet Inception Distance (FID) \cite{HeuselRUNH17} measures the divergence between generated and real image distributions; we compute it on a validation split of 50k triplets (sketch, reference, ground truth). MS-SSIM assesses structural similarity across multiple scales, and PSNR quantifies reconstruction fidelity via the decibel ratio to MSE. CLIP Score measures semantic alignment between the generated image and the ground truth via cosine similarity of their CLIP image embeddings. Unless otherwise stated, model selection and ablation conclusions are based primarily on FID.

The quantitative results are presented in Table \ref{table:quantitative}. Our method outperforms in FID, MS-SSIM, and CLIP score owing to the superior generalization and expressing ability of the network. Most existing methods may achieve acceptable results in lower resolutions but suffer from severe deterioration in perceptual quality in higher resolution, due to the ineffectiveness of synthesizing textures in a resolution much higher than the training dataset in $512^{2}$. MangaNinja achieves the best score in PSNR, with the proposed method ranks number 2. This is because the limited generation ability and resolution of MangaNinja prevents it from synthesizing complicated backgrounds, bright colors, and rich details of the figures. This close-to-average characteristic makes it advantageous in the calculation of PSNR. 

\begin{table}[t]
\vspace{-0.5em}
\setlength\tabcolsep{3pt}
    \centering
    \caption{Quantitative comparison 
    evaluated by 50K-FID, PSNR, MS-SSIM, and CLIP cosine similarity. \dag: reference images are randomly selected to be close to real-application scenarios and cover the corner cases. \ddag: References are deformed from ground truth. \S: Tested at $512^{2}$ resolution.}
    \vspace{-0.5em}
    \footnotesize{\begin{tabular}{|c|c|c|c|c|c|c|c|c|}
        \hline
        \multicolumn{5}{|c|}{Method} & {\dag FID $\downarrow$} & {\ddag PSNR$\uparrow$} & {\ddag MS-SSIM$\uparrow$} & {\ddag CLIP score$\uparrow$}\\
        \hline
        \multicolumn{5}{|c|}{Ours} & \textbf{8.28} & 28.83 & \textbf{0.70} & \textbf{0.912} \\
        \hline
        \multicolumn{5}{|c|}{Yan et al.~\cite{yan2025image}} &  12.09 &28.44 & 0.61 & 0.896\\
	\hline
        \multicolumn{5}{|c|}{ColorizeDiff~\cite{Yan_2025_WACV}} & 13.42 & 28.04 & 0.57  & 0.891 \\
        \hline
        \multicolumn{5}{|c|}{IP-Adapter-XL} & 36.61 & 28.23 & 0.44 & 0.758 \\
        \hline
	\multicolumn{5}{|c|}{IP-Adapter} &  94.53 & 27.94 & 0.50 & 0.762 \\
        \hline
        \multicolumn{5}{|c|}{T2I-Adapter} & 94.98 & 27.97 & 0.28 & 0.613  \\
        \hline
        \multicolumn{5}{|c|}{\S MangaNinja} & 42.85 & \textbf{29.64} &  0.67 & 0.892  \\
        \hline
    \end{tabular}}
    \label{table:quantitative}
\vspace{-1em}
\end{table}

\noindent\textbf{User study}
We employ a user study to demonstrate how the proposed method and existing methods are subjectively evaluated by individuals, where our method is tested against all compared methods. We prepare 25 image sets, for each image set, our method is compared to 6 other methods, and 4 comparisons between randomly chosen existing methods guarantees the reliability. 30 participants are involved, with 16 image sets shown to each individual. 

\begin{figure}[t]
    \centering
    \includegraphics[width=1\linewidth]{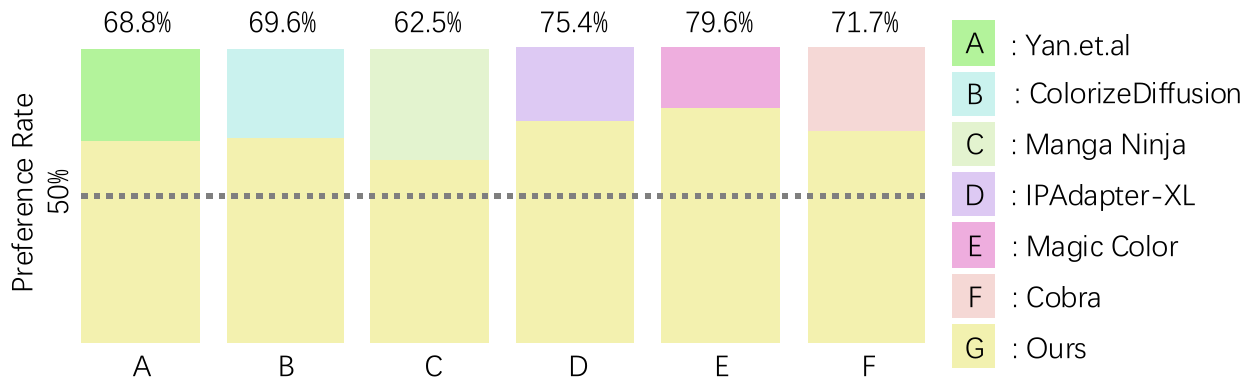}
    \vspace{-1em}
    \caption{Results of user study. Our method is preferred across all compared methods.}
    \label{user_study}
    \vspace{-1em}
\end{figure}

The results of the user study are illustrated in Figure \ref{user_study}, where the proposed method has received the most numbers of preferences among all the methods compared. Chi-Squared Test is applied to validate the comparison, where our method is preferred in all 6 comparisons and the results significantly differ from random selection at $p < 0.01$. All images shown in the user study are included in the supplementary materials.

\subsection{Cross-content validation}
We show the cross-content colorization results in Fig~\ref{fig:cross-content} to illustrate the ability of the proposed method to disentangle the spatial and style semantics, and to eliminate the spatial entanglement in severe cases. As shown in the Figure, when the sketches and references are from different domains (for example, characters and sceneries), the proposed method is still able to synthesize high quality results with pleasant visual quality, fine details, and free from artifacts. 
\section{Conclusion}
\label{sec:conclusion}
In this paper, we analyze the distribution shift problem for the image-referenced sketch colorization task and propose to model the distribution shift with a dual-branch architecture and optimize the problem with a Gram regularization loss. A tagger network trained on the anime-style dataset is integrated into the framework for fine-grained attribution control. Qualitative and quantitative experiments, together with a user study, validate our superiority over previous methods. Ablation study reveals the effectiveness of each module. We include the failure cases and discussions in the supplementary material. 


\section*{Acknowledgement}
We thank Chang Liu for providing the original sketch image data used in this work.

{
    \small
    \bibliographystyle{ieeenat_fullname}
    \bibliography{main}
}

\end{document}